\documentclass[letterpaper, 10 pt, conference]{ieeeconf}

\IEEEoverridecommandlockouts                              
\usepackage{amssymb}
\usepackage{graphicx,amsmath}
\usepackage{lineno}
\usepackage{caption,url}
\usepackage{float}
\usepackage{epstopdf}
\usepackage{authblk}
\usepackage{dblfloatfix}

\usepackage[
  separate-uncertainty = true,
  multi-part-units = repeat
]{siunitx}

\usepackage{subcaption}
\usepackage{gensymb}
\usepackage[colorlinks]{hyperref}
\hypersetup{
     colorlinks,
     linkcolor=black,
     citecolor=black,
     filecolor=black,
     urlcolor=blue,
 }

\begin{document}

\title{\LARGE \bf Learning Linear Policies for 
Robust Bipedal Locomotion \\ on Terrains with Varying Slopes 
}

\author{{Lokesh Krishna, Utkarsh A. Mishra, Guillermo A. Castillo, Ayonga Hereid, Shishir Kolathaya}

\thanks{This work is supported by the Robert Bosch Centre of Cyber Physical Systems, Pratiksha Trust and OSU under the M\&MS Discovery Theme Initiative.}
\thanks{L. Krishna is with the department of Electronics Engineering, Indian Institute of Technology (BHU) Varanasi, India}
\thanks{U. A. Mishra is with the Mechanical and Industrial Engineering Department, Indian Institute of Technology Roorkee, India}
\thanks{G. A. Castillo and A. Hereid are with the department of Mechanical and Aerospace Engineering, Ohio State University, Columbus, OH, USA}
\thanks{S. Kolathaya is with the department of Computer Science and Automation and the Centre for Cyber-Physical Systems, Indian Institute of Science, Bengaluru, India}
}

\maketitle

\thispagestyle{empty}
\pagestyle{empty}

\begin{abstract}

In this paper, with a view toward deployment of  light-weight control frameworks for bipedal walking robots, we realize end-foot trajectories that are shaped by a single linear feedback policy. We learn this policy via a model-free and a gradient free learning algorithm, Augmented Random Search (ARS), in the two robot platforms Rabbit and Digit. Our contributions are two-fold: a) By using torso and support plane orientation as inputs, we achieve robust walking on slopes of upto 20$\degree$ in simulation. b) We demonstrate additional behaviors like walking backwards, stepping-in-place, and recovery from external pushes of upto 120~N. The end-result is a robust and a fast feedback control law for bipedal walking on terrains with varying slopes. Towards the end, we also provide preliminary results of hardware transfer to Digit. 



\end{abstract}







\section{Introduction}

 Locomotion for legged robots has been an active field of research for the past decade owing to the rapid progress in actuator and sensing modules. Actuators like the BLDC motors have become efficient and powerful, and sensors like the IMUs have become more accurate and affordable. As a result, there has been a spurt in the growth of high performance legged robots over the last few years \cite{Bledt2018MIT,Leeeabc5986}. In particular, there has been rapid progress in the domain of quadrupedal robots, and there is a general push for the realization of natural and efficient locomotion behaviours comparable with animals. For example, the MIT Cheetah uses a combination of Model Predictive Control (MPC) and whole-body impulse control, enabling it to walk on rough terrains and generate back-flips \cite{Bledt2018MIT}. Similar controllers are in use for Spot from Boston Dynamics, Vision 60 from Ghost Robotics and others. However, these controllers, despite being successful, are using a simplified model (the ``single potato" model), which is not applicable for bipeds. Hence, the controller development for bipeds and quadrupeds have largely had their own separate paths. 


\begin{figure}[!t]
\centering
\vspace{2mm}

\includegraphics[width =0.9\linewidth]{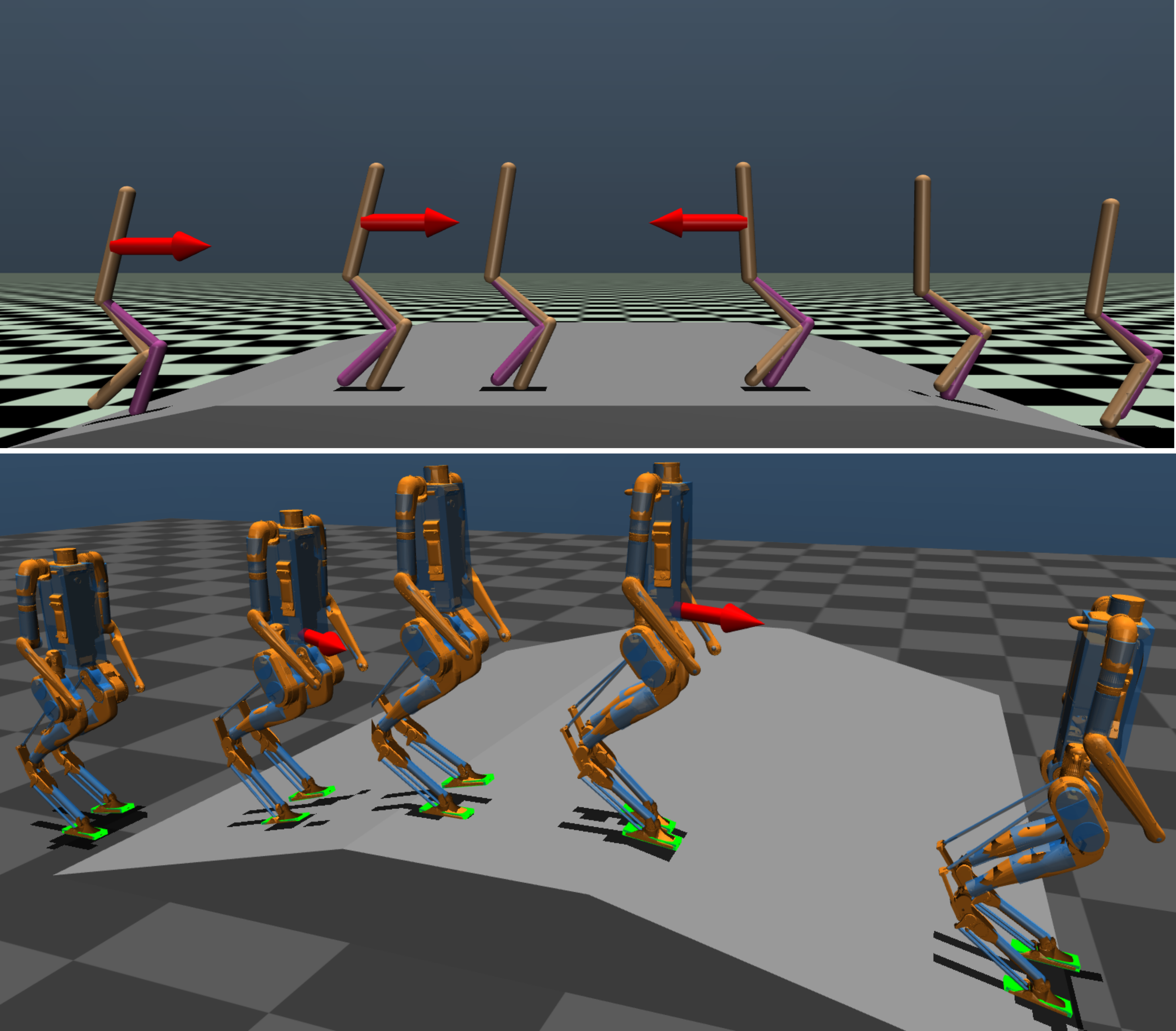}
\caption{Figure showing Rabbit (Top) and Digit (Bottom) in Simulation: Traversing incline and decline with robustness towards adversarial forces}
\label{fig:intro_train}
\vspace{-5mm}
\end{figure}

While numerous works have demonstrated successful bipedal controllers, effective bipedal locomotion strategies which accomplish robust walking on rough terrains are heavily engineered and layered, consisting of several concepts being stacked to get a stable walking controller. For example, SLIP models in conjunction with Raibert's controller are used to achieve stable walking \cite{apgar2018rss}.
Other methods include the centroidal models, the Zero-Moment Point (ZMP) \cite{tedrake2004}, the linear inverted pendulum \cite{gong2020angular}, and Hybrid Zero Dynamics (HZD) \cite{hzd_grizzle}. They are usually combined with either offline or online trajectory optimization to obtain a library of behaviors \cite{heried2018cfrost,xiong2018ames}. While the offline optimization uses the full body model, the online part uses an approximate model in order to solve it in real-time. 

Apart from the classical approaches that are heavily influenced by model based design, Reinforcement Learning (RL) based approaches have gained popularity today \cite{guillermo3D,siekmann2020learning,2020-ALLSTEPS,deeploco}. 
Physics-based bipedal character animation \cite{deeploco} has shown benefits of using Deep Reinforcement Learning (DRL) algorithms by generating robust locomotion strategies for a diverse set of terrains. \cite{cdmloco} learned centroidal dynamics based motion planners for multi-legged systems in rough terrain by using DRL stacked with their inverted pendulum on cart (IPC) model. Similarly \cite{siekmann2020learning}, \cite{cassie2019sim2real} use novel DRL frameworks for walking on flat terrain and present a direct policy transfer of learned strategies from simulation to a physical robot.
%
Also, some of the works like
\cite{guillermo3D,da2019grizzle} have focused on blending the classical and learning based approaches. 
However, all of the approaches mentioned are computationally intensive, requiring several thousands of parameters for the neural network policy. In this paper, we would like to develop a control framework that neither uses extensive dynamic modeling, nor uses simplifying assumptions (like in quadrupeds). More importantly, we would like to use control policies that are computationally less intensive.


\subsection{Contributions and Paper Organization}

Our primary focus in this paper is walking on varying slopes and elevated terrains, which bring in a new set of challenges. We would like to follow a methodology similar to \cite{Leeeabc5986}, which uses a high-level policy that shifts the end-foot trajectories to realize robust walking on a diverse set of terrains. In particular, we propose a  formulation for bipedal locomotion that generates linear control policies, that not only transform the semi-elliptical trajectories, but also reshape them.
This was used to realize robust walking in the quadrupedal robot, Stoch 2, in simulation \cite{paigwar2020robust}. Based on the support plane and torso orientation, appropriate trajectories are generated,
which are then tracked via a joint level PD control law. With a structured low-level tracking controller, the policy is freed from learning the joint level coordination and non-linear control relations, thus making our approach intuitive. 
The contributions of the paper are as follows:

\textbf{Low computational overhead:} The main contribution of this paper is on training robust bipedal walking on challenging terrains with low computational footprint. By incorporating the priors of bipedal walking strategies and formulating a relevant control structure,  we significantly decrease the number of learnable parameters to  9 for Rabbit (2D) and 80 for Digit (3D). Our linear control policies are considerably compact compared to even the smallest of the neural network policies \cite{guillermo3D} capable of 3D bipedal walking. We firmly believe that having such a simplistic linear policy will greatly leverage our policy's explainability compared to any of the previous works,  \cite{cassie2019sim2real}, \cite{siekmann2020learning}. This feature allows us to obtain intuitive seed policies that enhance the sample efficiency of learning and realization of practical walking behaviours.

\textbf{Unified framework:} We previously achieved robust locomotion via linear policies in quadrupeds \cite{paigwar2020robust}.  Given the success behind this policy, we focused on implementing this methodology in bipeds, which are much harder to control. More importantly, bipeds have feet, which play a major role in stabilizing the torso. We have a unified learning-based control framework deployed alike for quadrupeds and bipeds. 

\textbf{Robustness and generalization:} We demonstrate walking on slopes of up to $20\degree$, walking backwards, and with feet. Our controller can reject disturbances, adapt to terrain transitions, and recover from adversarial force perturbations up to $120$ N. We validate our results on two bipedal robots, Rabbit and Digit, in simulation and show preliminary transfer of the policies in Digit hardware.

The paper is structured as follows: Section \ref{sec:background} will provide the description of the robot test-beds and the associated hardware considerations. Section \ref{sec:approach} will describe the control framework for the proposed strategy, followed by the description of the training process used in Section \ref{sec:training_eval}. Finally, the simulation results will be discussed in Section \ref{sec:simulation_results}.


\section{Robot model and Hardware testbed}
\label{sec:background}

In this section, we will briefly describe the robots used to demonstrate the proposed methodology. Specifically, we will describe the two bipeds, Rabbit and Digit, and the associated actuator-sensor framework that we aim to use for walking.



 


\begin{figure*}[!t]
    \centering
\includegraphics[width=0.6\columnwidth]{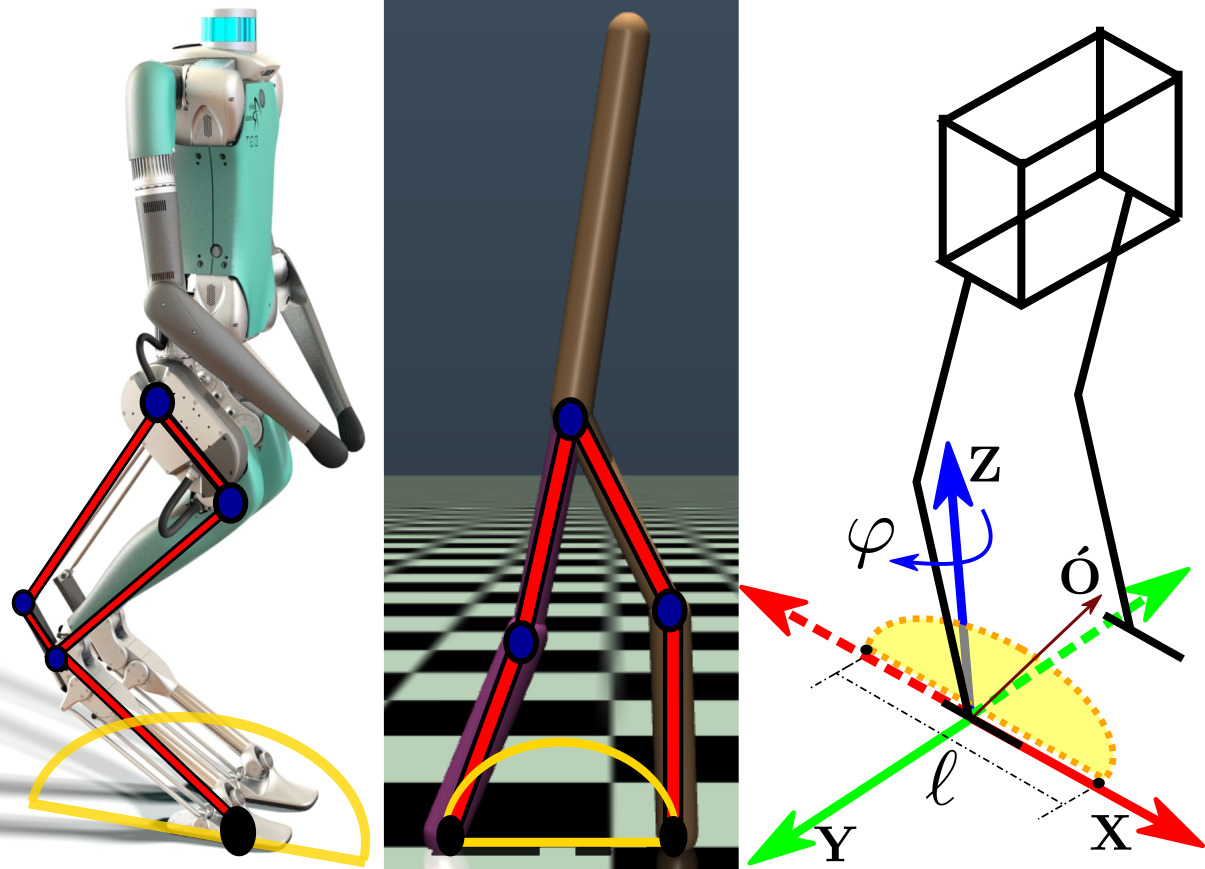}
    \includegraphics[width=1.4\columnwidth]{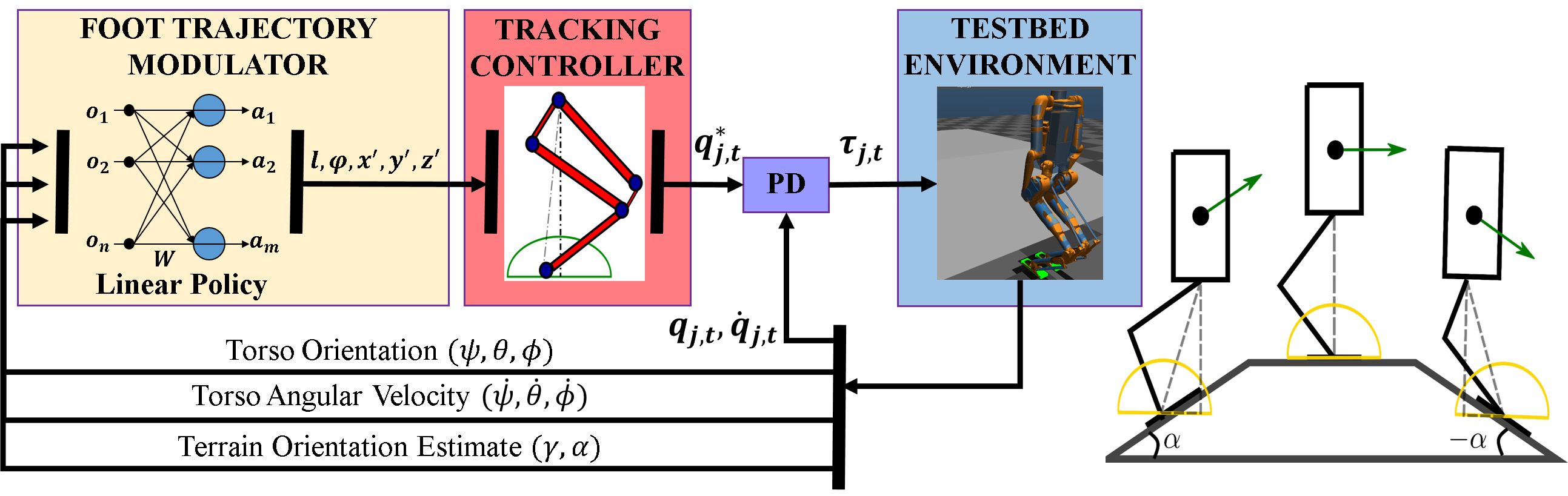}

    \caption{Figure showing the parameterization of the semi-elliptic trajectory for the feet (left). Figure showing the overall control pipeline (center) and the required shifting strategies of the semi ellipses for walking on slopes (right).}
    \label{fig:control_arch}
    \vspace{-5mm}
\end{figure*}



\textbf{Rabbit} \cite{rabbit} is a planar (2D) bipedal robot having $5$ rigid links and $4$ actuators. The links are torso, thighs and shins. 
The simulated model of Rabbit 
weighs $32$ kg with $12$ kg on the torso. Each leg weighs $10$ kg divided into $6.8$ kg and $3.2$ kg for the thigh and shin respectively. The legs form a 2R planar linkage, thus making the forward and inverse kinematics straightforward.

\textbf{Digit} is a $30$ DOF 3D biped developed by Agility Robotics, USA (see Fig. \ref{fig:control_arch} ahead). 
The total weight of the robot is $48$ kg, with approximately $22$ kg for the upper body and $13$ kg for each leg. 
Each arm has $4$-DOF for basic manipulation tasks. In this paper, we do not utilize the arm joints for balancing, hence, they will be controlled at fixed angles. Each leg consists of eight joints, including three actuated hip joints and one actuated knee joint. 
Two specially designed leaf-spring four-bar linkages are used to control two passive joints to provide additional compliance. The ankle roll and pitch joints are collectively controlled via two four-bar linkages to improve balancing and stability on a wide variety of surfaces. 

\subsubsection{Inverse Kinematics of Digit}
To solve the inverse kinematics (IK) for the special closed chains structure,
we keep the yaw hip angle constant and use simple trigonometric computations to transform the foot Cartesian position into virtual leg length, pitch angle and roll angle, where the virtual leg is the imaginary line that connects the hip of the robot with the leg ankle. In this decoupled system, the virtual leg length and pitch angle are determined by the position of the hip pitch and knee joint. With these values, we then solve the ``reduced" IK subject to the constraints imposed by the leg kinematic structure. Finally, by solving the IK problem for a sufficiently diverse set of desired foot positions, we obtain a closed form solution using nonlinear regression. 

\subsubsection{Ankle Regulation}

Due to the significant foot size and additional degrees of actuation in ankle, we provide an explicit ankle regulation to align the feet parallel to the terrain. Also, to have a smooth transition of the swing leg during contact, we further reduce the ankle torques to about $7.5\%$ of the nominal values to make joints nearly passive.










\section{Methodology}
\label{sec:approach}

In this section, we provide an overview of the control framework, i.e., we describe how the end-foot trajectories are shaped and tracked in real-time that result in walking on slopes. This framework is also shown pictorially in Fig. \ref{fig:control_arch}.



\subsection{Controller Structure}
\label{subsec:controller_structure}
We use a hierarchical structure, as shown in Fig. \ref{fig:control_arch} (center), that decouples the linear and nonlinear parts of the control framework. The linear part is the foot trajectory modulator, and the nonlinear part is the tracking controller.
Based on the orientation of torso and supporting plane, the modulator transforms the semi-ellipses in real-time. 
The tracking controller then tracks the generated semi-ellipses (refer \ref{subsec:ellipse_formulation}) independently for each leg with a constant phase difference\footnote{It is worth noting that by varying this phase difference, we can obtain different types of gaits like running, skipping, and hopping.}. Joint-level PD controllers are used to track the required joint targets obtained through inverse kinematics. Note that our controller does not include contact modelling and explicit switching conditions \cite{hzd_grizzle}, as a model-free learning algorithm is used.

\subsection{Reinforcement Learning Formulation}
\label{subsec:obs_action}

Given the state space $\mathcal{S}\subset \mathbb{R}^n$ of dimension $n$, and action space $\mathcal{A}\subset \mathbb{R}^m$ of dimension $m$, we formulate our linear policy $\pi: \mathcal{S}\to \mathcal{A}$ as $\pi(s):= Ms$, where $M\in \mathbb{R}^{m\times n}$ is a matrix consisting of the parameters to be learnt.
For this policy, we detail the observation and the action space below.


\textbf{Observation Space:} While various works \cite{siekmann2020learning} demonstrated the usage of full robot state with neural networks, \cite{paigwar2020robust} shows a simplification by selecting suitable observations. However, the direct transfer of \cite{paigwar2020robust} to bipedal robots was unsuccessful as it fails to capture the sudden change in torso orientation, which is crucial to maintain balance during terrain transitions. The torso's angular velocity as feedback made the policy compact in size and more sensitive to external disturbance.  
Thus, the observation space is a 8 dimensional state vector defined as $s_t$ = $\{\psi, \theta, \phi, \dot{\psi},  \dot{\theta}, \dot{\phi}, \gamma, \alpha \}$, where $ \psi,\theta,\phi$ are the roll, pitch and yaw of the torso with $\dot{\psi}, \dot{\theta}, \dot{\phi}$ being the corresponding angular velocity terms, and $\gamma, \alpha$ represent the support plane roll and pitch, respectively.


\textbf{Action Space:} The actions represent the parameters for the semi-ellipse i.e., parameters include the length of the major axis $(\text{Step Length} - \ell)$, the orientation of the ellipse along the Z-axis $(\text{Steering Angle} - \varphi)$  and translation along X, Y and Z through independent shifts $\acute{x},\acute{y},\acute{z}$ (together shown as $\acute{O}$). The above transformations are defined in the local leg-frame of reference, as seen in  Fig. \ref{fig:control_arch} (left). These actions enable the policy to learn an optimal strategy based on the nature of the underlying terrain as shown in Fig \ref{fig:control_arch} (right). We also extend the step lengths to have negative values and allow the policy to take backward steps when the torso is pushed back. This ensures smooth transition from forward to backward walking and enables the robot to reorient its torso while falling backwards. The action space is a $10$-dimensional vector with each leg having: $\{\ell,\varphi,\acute{x},\acute{y},\acute{z}\}$.

\subsection{Semi-Elliptical Trajectory Generation}
\label{subsec:ellipse_formulation}
Taking the hip as the origin for each leg, we generate a planar ellipse as follows with $x,y,z$ being the position of the foot as:
\begin{gather}
x = \frac{l}{2}\cos{(2\pi(1-\zeta))}, \quad y = 0 \quad \quad \zeta \in [0, 1)\nonumber \\
z=\begin{cases}
          -h_d &\zeta \in [0, 0.5), \text{stance}. \\
          -h_d + h_{sf}\sin{(2\pi(1-\zeta))} &\zeta \in [0.5, 1), \text{swing}. \\
     \end{cases}\nonumber
\end{gather}
where $h_d$ is the desired hip height from the support plane, $h_{sf}$ is the maximum height of the swing foot (minor axis of the ellipse) and $\zeta$ is the phase of the gait cycle.
The semi-elliptical trajectory which lies in the $X$-$Z$ plane is then transformed according to the action by:
\begin{align}
    \bar{x} = \acute{x} + x \cos(\varphi), \quad
    \bar{y} = \acute{y} + x \sin(\varphi), \quad
    \bar{z} = \acute{z} + z,\nonumber
\end{align}
where $\bar{x}$, $\bar{y}$ and $\bar{z}$ are the coordinates of the foot in the leg frame of reference after the transformation.

\section{Policy Training and Evaluation}
\label{sec:training_eval}
 In this section, we describe the training and evaluation procedure used for successfully learning the linear policy. 
 In order to improve sample efficiency and ensure convergence to walking policies, we first develop an initial policy based on existing bipedal control strategies. The removal of such an initial heuristic seed policy and opting to learn from scratch often lead to the failure of the policies.  



\subsection{Development of Initial Policies}
\label{subsec:dev_IP}

We actively vary the step length of the robot's gait to maintain balance in the sagittal plane. This is realized by relating the torso pitch and the angular velocity about that axis with the step length, as shown below.

\begin{equation}
\label{equation:1}
    \ell = \ell_{\circ}+ K_{\ell}(\theta - \theta_{d}) + K_{\dot{\ell}}\dot{\theta},
\end{equation}
where $\ell_{\circ}$, is the desired step length when the torso pitch error is driven to zero. This prevents the training from converging to a sub optimal ``walking in place" policy. Similarly, in the following equation, we extend this strategy to the frontal plane, relating the torso roll components with the y-shifts:

\begin{equation}
\label{equation:2}
    \acute{y} = K_{\acute{y}}(\psi - \psi_{d}) + K_{\dot{\acute{y}}}\dot{\psi}.
\end{equation}

Since we aim to obtain policies that walk with a constant heading direction, the semi ellipses must be steered to correct the drift in the torso's yaw, giving rise to the following:
\begin{equation}
\label{equation:3}
    \varphi = K_{\varphi}(\phi - \phi_{d})+ K_{\dot{\varphi}}\dot{\phi}.
\end{equation}
The above three feedback laws are sufficient to obtain good initial seed policies for walking on flat ground. To extend this further for elevated terrains, we introduce $\acute{x}$, to shift the semi ellipses along the slope and $\acute{z}$, to maintain the torso height, which gets disturbed upon the application of an x-shift. 

\begin{align}
\label{equation:4}
    \acute{x} = K_{\acute{x}}(\theta - \theta_{d}  - \alpha ) + K_{\dot{\acute{x}}}\dot{\theta}, \quad 
    \acute{z}=K_{\acute{z}}\alpha,
\end{align}
where, $K_{\square}$ (with the square subscript being the respective state term) is the gain relating the state variable with the corresponding action, $\psi_{d}, \theta_{d}, \phi_{d},$ are the desirable torso roll, pitch, yaw. The above equations enables us to encode the desired behavior directly in the parameter space and allows the learning algorithm to further enhance the policy’s performance along the training. The policy matrix parameters could be directly interpreted as the gains discussed above. To obtain a good initial policy, we manually tune these gains by limiting the maximum allowed error in the defined state variables and mapping them to the clipped action space values based on the robot's kinematic limits. For example, we expect the policy to have the maximum step length when the torso pitch disturbance is at a certain maximum value. 

The seed policy is then obtained by initializing the corresponding parameters in the sparse policy matrix. It is worth mentioning that the equations discussed above are just to get a good initial guess and this structure is not imposed throughout the training process. These initial policies are sub-optimal in nature, and often fail to generalize across different slopes, which is later obtained through training.  

\subsection{Training Details}

\textbf{Training algorithm:} With the initial policies described previously, we use Augmented Random Search (ARS) \cite{mania2018simple}, a model-free and gradient-free learning algorithm to obtain an optimal linear policy.
%
%
We preferred ARS as it is easy to use, has fewer hyperparameters to tune and directly optimizes a deterministic policy using the method of finite-differences.

\textbf{Reward Function:} A reward function similar to \cite{paigwar2020robust} is used with two major changes. $(i)$ Addition of a velocity reward term to prevent unrealistic walking speeds. This update was significant as the absence of this term lead to severe drifting and leaping motions of the policy. $(ii)$  The removal of the standing penalty term, as we start the training with a walking policy. We have
\label{sec:customreward}
\begin{align}
r= & \: G_{w_1}(\psi)+G_{w_2}(\theta)+G_{w_3}(\phi) +G_{w_4}(h-h_d) \nonumber \\
 & + G_{w_5}(v_{x}- v_{nx})+ W \Delta b_{x},
\end{align}
where, $v_{x}, v_{nx}$ are the current and nominal torso linear velocity along the x-direction; $h$ and $h_{d}$ are the current and desired height of the robot's COM, and $\Delta b_{x}$ is the distance travelled by the COM along the heading direction weighted by $W$. The mapping $G: \mathbb{R} \to [0, 1]$ is the Gaussian kernel function, and is given by $G_{w_j}(x)=\exp{(-w_j*x^2)}, \: w_j > 0$.

\textbf{Training Setup:} As seen in Fig. \ref{fig:intro_train}, we use custom tracks that are divided into incline, decline and flat plateau regions. A random track was uniformly sampled  from a discrete set of inclinations and the robot is deployed in different locations along these tracks while training. The tracks were designed with a significant plateau region to train the policy to adapt to the terrain transitions. Every episode is terminated when the episode length is reached or if the robot’s torso is below a certain height or about to topple. The hyperparameters of ARS used in training are: learning rate $(\beta)$~=~0.03, noise~$(\nu)$~=~0.04 and episode length~=~10k simulation steps.

\section{Simulation Results}
\label{sec:simulation_results}


\begin{figure*}[!t]
    \centering
      \includegraphics[width=0.695\columnwidth]{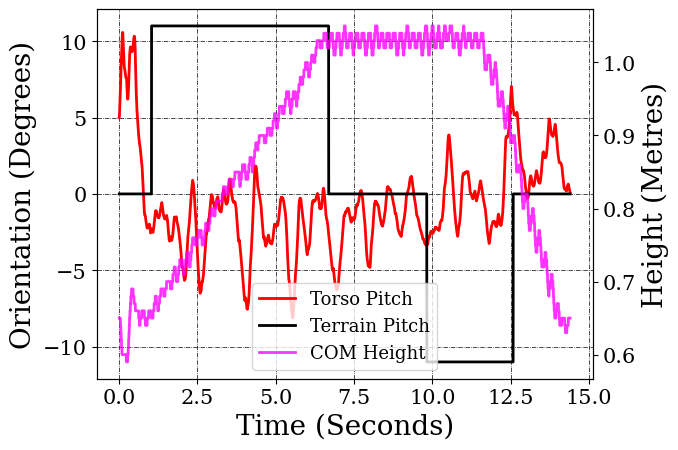}%
      \includegraphics[width=0.670\columnwidth]{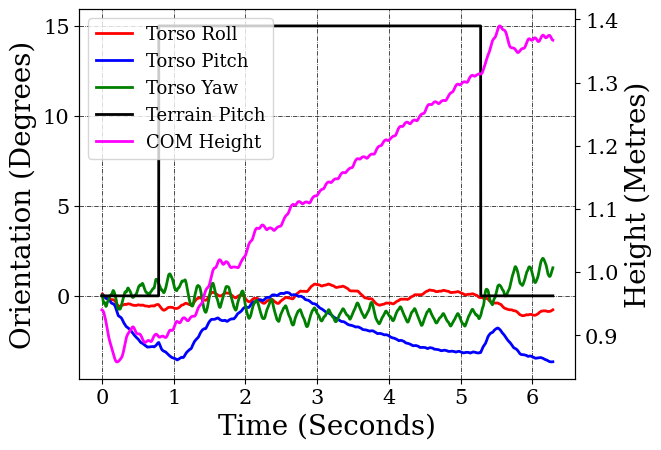}%
      \includegraphics[width=0.705\columnwidth]{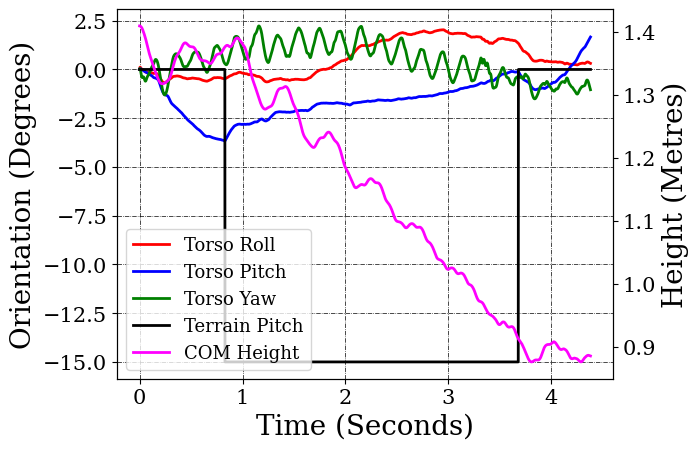}%

    \caption{Figures showing Rabbit walking on a $11^\circ$ track with incline and decline (left). Torso profile of Digit while walking on $15^\circ$ incline (centre) and decline (right).} 
    \label{fig:on_slopes}%

\end{figure*}
\begin{figure*}[!t]
    \centering
      \includegraphics[width=0.68\columnwidth]{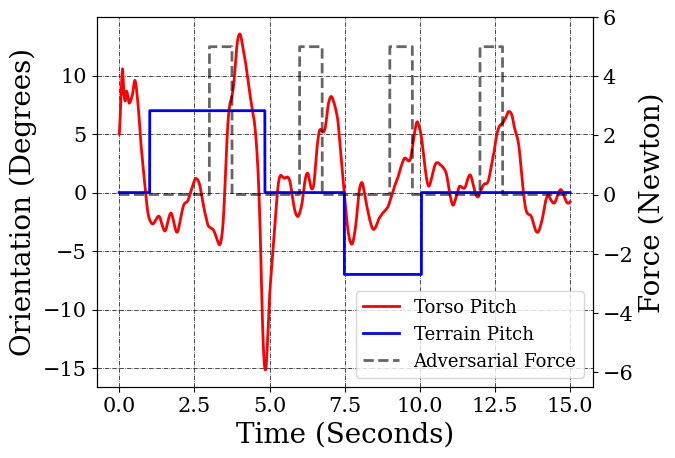}%
      \includegraphics[width=0.68\columnwidth]{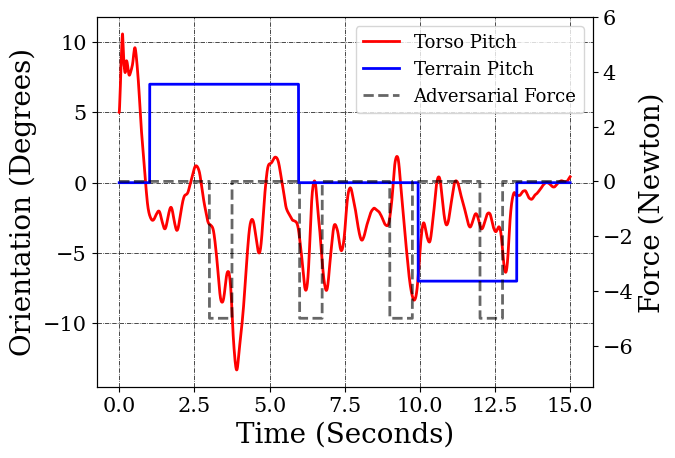}%
      \includegraphics[width=0.68\columnwidth]{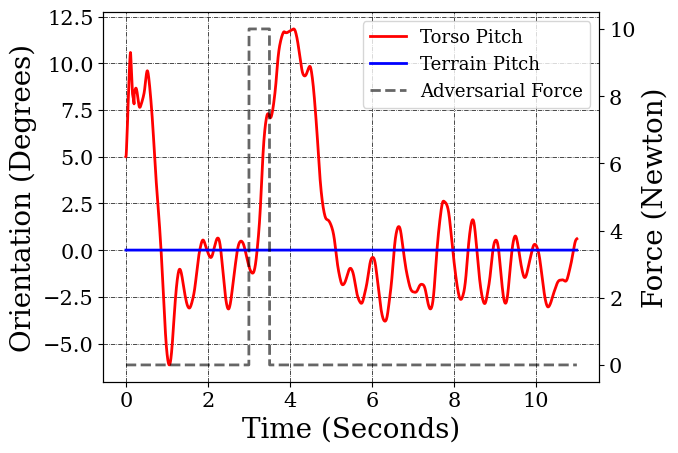}%
    \caption{Figures showing disturbance rejection in the torso pitch upon the application of perpendicular forces $5$~N(left), $-5$~N (centre) directly to the torso of Rabbit while walking on a $7^\circ$ track and a $10$~N impulsive force applied while walking on flat ground.}
    \label{fig:disturbance_rejection_rabbit}%

\end{figure*}
\begin{figure*}[!t]
    \centering
      \includegraphics[width=0.67\columnwidth, height=0.5\columnwidth]{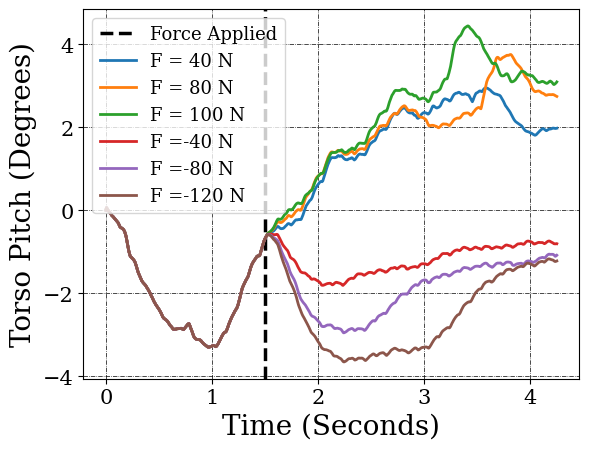}%
      \includegraphics[width=0.67\columnwidth, height=0.5\columnwidth]{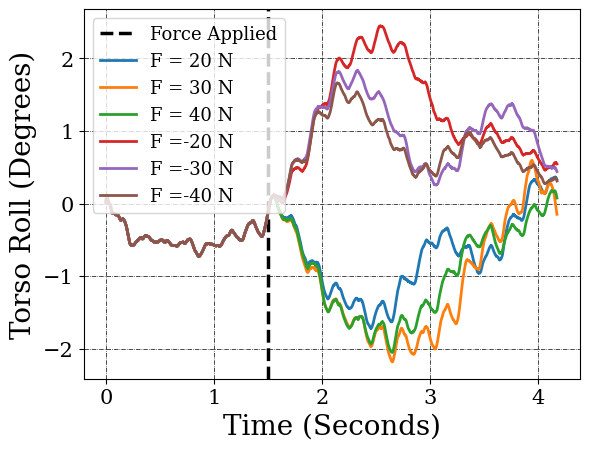}%
      \hspace{0.5cm}
      \includegraphics[width=0.55\columnwidth, height=0.5\columnwidth]{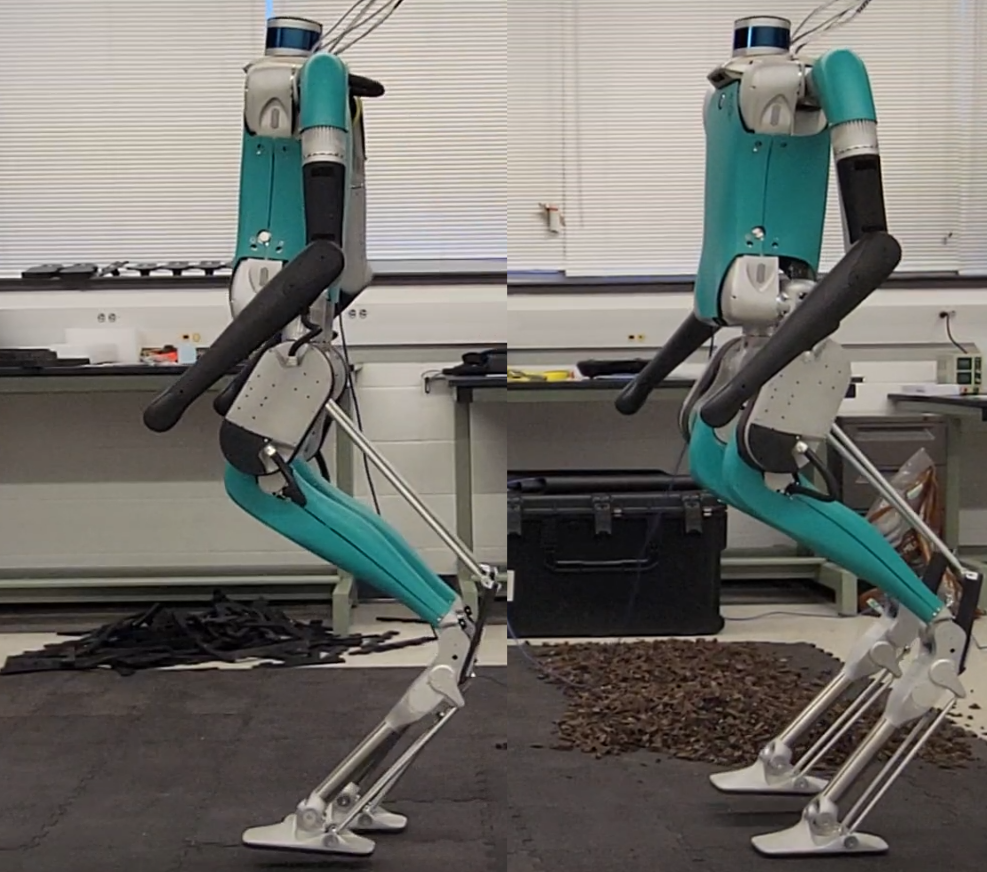}

    \caption{Figures showing the disturbance rejection in torso pitch(left) and roll(center) upon the application of adversarial forces on Digit along the longitudinal and lateral axis respectively. The figure on the right shows the experimental setup where we performed preliminary transfer of the policy from simulation to hardware. 
    } 
    \label{fig:disturbance_rejection_digit}%
\vspace{-5mm}
\end{figure*}

We use the MuJoCo physics engine to validate our proposed method. A custom Gym environment was developed for the training and testing of our policies.

Based on Rabbit's simple model (planar with $5$ links) and the legs' symmetric configuration, we made the policy learn a single set of semi-elliptical parameters and tracked it in an out-of-phase manner for both the legs. With the reduced state space 
$(\theta,\dot{\theta},\alpha)$ and action space  $(\ell,\acute{x},\acute{z})$, we train a compact linear policy 
capable of planar walking on flat ground, incline, decline and handle adversarial force perturbations.
On the other hand, Digit requires independent semi-ellipses for each leg to realize stable 3D walking. This is because our formulation requires asymmetric trajectories to maintain the torso orientation along the frontal plane. 

We also perform Forward Kinematics (FK) based terrain estimation for Rabbit using the surface tangent vector connecting two successive foot contact points. However, for Digit we forgo the FK based terrain estimation due to the availability of sophisticated perception sensors on board. We obtain the terrain elevation from the simulation to orient the foot along the support plane.

This section presents the performance analysis of the control policy in terms of $(i)$ Generalization towards multiple slopes and $(ii)$ Disturbance rejection towards external perturbations observed in both the robots.          

\subsection{Uphill and Downhill walking}

The concerned robots were trained on slopes with varying incline and decline angles sampled from $\{7\degree, 11\degree, 13\degree\}$. Walking on flat terrain was inherently achieved without being explicitly trained. Measurements collected during the simulation experiments for Rabbit and Digit are shown in Fig. \ref{fig:on_slopes} where they traverse a $11\degree$ and a $15\degree$ slope, respectively. 


\textbf{Rabbit} motion in Fig. \ref{fig:on_slopes} (left) depicts torso regulation during all three terrain cases and treats terrain transition as a disturbance in torso. The COM height exhibits a natural behavior as the robot takes a longer time to cover up-slope and walks relatively faster down-slope.


\textbf{Digit} motion in Fig. \ref{fig:on_slopes} (center, right) illustrates the torso behaviour while moving up and down a slope. The torso shows minimum and controlled oscillations (between $\pm3\degree$), which is inherent to the walking motion. As the robot has a 6-DOF floating base, the torso regulations are relatively difficult as compared to Rabbit. By principle, to balance itself, the torso should be rolled towards the support region (stance foot) to deploy the swing phase effectively. Such emergent behaviour is seen in our policy, as depicted in Fig 4.   The policy can recover from the torso orientation disturbance caused during terrain transitions(as seen at $t = 0.8, 5.2$ secs in Fig \ref{fig:on_slopes} (centre) and $t = 0.9, 3.7$ secs in Fig \ref{fig:on_slopes} (right)). The torso is also maintained at a constant height from the support plane irrespective of the terrain elevation and hence shows a consistent increase and decrease in incline and decline, respectively. It is worth noting that our policies showed zero-shot generalizations towards higher elevations that were unseen during training. Thus by training for $-13\degree$ to $13\degree$, we obtained policies that could traverse higher slopes ranging from $-15\degree$ to $20\degree$ as seen in the attached video (refer \ref{sec:conclusion}). 

\subsection{Disturbance Rejection and Robustness}

The proposed approach comes up with inherently stable and robust policies. As our elliptical trajectories are derived from fixed reference ellipse as shown in Fig. \ref{fig:control_arch} (right), the relative shifts and steering angle change the time of collision with the terrain. Hence, the dynamic contact interactions with the terrain are themselves not periodic and treated as implicit disturbances. The resulting final policies were validated using impulses and periodic forces. 

\textbf{Rabbit:} Compared with the most recent robustness baseline for Rabbit as demonstrated in \cite{guillermo2D}, we conduct an equivalent robustness test to show the strength of kinematic planning with linear policies against HZD planning with neural network policies. 
The adversarial forces applied and the torso correction behavior on different phases of $7\degree$ slopes are shown in Fig. \ref{fig:disturbance_rejection_rabbit} (left and center). 
When forward and backward periodic forces of $5$ N magnitude was applied for $0.25$ secs, Rabbit self-corrected by taking back-steps. A related illustration can be visualized from Fig. \ref{fig:intro_train} (top) and is shown in the attached video. 
As an impulse reaction test, a force of $10$ N was applied during flat ground walking, and the torso balancing is demonstrated with the help of Fig. \ref{fig:disturbance_rejection_rabbit} (right), where Rabbit was able to recover from disturbances as high as $\sim 12\degree$.

\textbf{Digit:} We present our controller's robustness even with significant torso inertia and demonstrate a strong baseline for Digit. For comparison, we consider the tests done by \cite{guillermo3D} for Cassie, which is a predecessor to Digit. 
We apply longitudinal forces ranging from $-120$ N to $100$ N and lateral forces ranging from $-40$ N to $40$ N to the floating base respectively. The self-correction behaviors are demonstrated in Fig. \ref{fig:disturbance_rejection_digit}. After applying a force at time $t = 1.5$~s that last for a significant period ($0.1 - 0.3$ \text{secs}), the pitch and roll corrections were observed. In Fig.~\ref{fig:disturbance_rejection_digit} (left), the policy was able to converge to a minimal pitch deviation of $\sim 2\degree$ for forward forces and deviations of $\sim 0.5\degree$ for backward pushes of magnitude as high as $120$ N.
The lateral forces contributed to the disturbances in torso roll which resulted in significant y-shifts towards the roll direction in order to balance and correct it back to $0\degree$. The corresponding regulation behavior is shown in Fig. \ref{fig:disturbance_rejection_digit} (middle) where the policy managed to decrease the error in torso roll to less than $\sim 0.5\degree$. Thus, the proposed strategy successfully recovered from the loss of balance and stability caused due to unknown disturbances. 


\section{Conclusion}

\label{sec:conclusion}



In this paper, we successfully demonstrated the development of a single linear policy capable of robust bipedal locomotion on terrains with varying slopes for two robots, Rabbit and Digit. The end-foot trajectory modulating policy is shown to generalise across inclines, declines and terrain transitions. Without the need for explicit training, our approach produces inherently robust policies that recover from significant force perturbations. The current approach, along with our previous work \cite{paigwar2020robust}, provide a unified framework that can quickly synthesize feedback control policies for multi-legged robots, thus radically simplifying the process of controller design and implementation for rough terrain walking. The linear policies are highly compute-efficient and the smallest in size compared to other learning-based policies. We also show preliminary results on Digit hardware, Fig. \ref{fig:disturbance_rejection_digit} (right). 
Future work will involve implementing walking on slopes in hardware, investigating the controller's capabilities to its lees and extending it to challenging terrains. 
The video presentation accompanying this paper is shown here: 
\textbf{\href{https://youtu.be/4WhgD8u74OY}{https://youtu.be/4WhgD8u74OY}}.










\bibliographystyle{ieeetr}
\bibliography{references}

\end{document}